\documentclass{optica-article}

\journal{opticajournal} 

\articletype{Research Article}

\usepackage{lineno}

\usepackage{wrapfig,lipsum,booktabs}
\usepackage{multirow}
\usepackage{siunitx}
\usepackage{float}

\newcommand{\RNum}[1]{\uppercase\expandafter{\romannumeral #1\relax}}

\newcommand{\cmt}[4]{\ifx\DRAFT\undefined\else\colorbox{#3}{\textcolor{#4}{\small{\textsf{[\textbf{#1}: #2]}}}}\fi}

\begin{document}

\title{Video-based sympathetic arousal assessment via peripheral blood flow estimation}

\author{Björn Braun,\authormark{1} Daniel McDuff,\authormark{2} Tadas Baltrusaitis,\authormark{3} and Christian Holz\authormark{1,*}}

\address{\authormark{1}Department of Computer Science, ETH Zürich, Switzerland\\
\authormark{2}University of Washington\\
\authormark{3}Mesh Labs, Microsoft}

\email{\authormark{*}christian.holz@inf.ethz.ch}

\begin{abstract*} 
Electrodermal activity (EDA) is considered a standard marker of sympathetic activity. 
However, traditional EDA measurement requires electrodes in steady contact with the skin. 
Can sympathetic arousal be measured using only an optical sensor, such as an RGB camera? 
This paper presents a novel approach to infer sympathetic arousal by measuring the peripheral blood flow on the face or hand optically. 
We contribute a self-recorded dataset of 21 participants, comprising synchronized videos of participants’ faces and palms and gold-standard EDA and photoplethysmography (PPG) signals.
Our results show that we can measure peripheral sympathetic responses that closely correlate with the ground truth EDA.
We obtain median correlations of 0.57 to 0.63 between our inferred signals and the ground truth EDA using only videos of the participants’ palms or foreheads or PPG signals from the foreheads or fingers. 
We also show that sympathetic arousal is best inferred from the forehead, finger, or palm.

\end{abstract*}

\section{Introduction}
\label{sec:introduction}
The human body regulates states of physiological arousal and responds to stressors via innervation of the autonomic nervous system (ANS). 
Sympathetic arousal, often characterized as the ``fight or flight response'', elevates heart rate and blood pressure, dilates sweat glands, and is commonly measured using heart rate variability (HRV) or electrodermal activity (EDA).
Recent evidence suggests that a number of often used metrics from the heart rate variability (HRV) may not provide a very reliable measure of sympathetic activity~\cite{billman2013edasymp, thomas2019edasymp, mcduff2020peripheralhemo}.
Electrodermal activity (EDA), on the other hand, is a frequently employed~\cite{dawson2016handbookeda, posada-quintero2020edareview} and standardized psychophysiological measure~\cite{brown1967proposed} of sympathetic arousal that captures the dilation of the sweat glands and is considered to be an exclusively sympathetic marker~\cite{boucsein2012eda, critchley2002edasympathetic}.
EDA measurement, with limited exceptions~\cite{pavlidis2012perinasalsweat}, requires contact with the skin via conductive electrodes strapped to sites on the body with a high density of sweat glands, such as the fingers, palms, and feet~\cite{marieke2012sweatinglocations, zeagler2017sweatinglocations}.  
Most recently, smartwatches with electrodermal sensors placed on the wrist (such as the Empatica E4) have gained popularity as a wristworn sensor is much more practical in everyday life. Despite the substantial advancements in wearable sensing over the last few years, certain limitations remain.
EDA sensors are far from ubiquitous, the majority of smartwatches do not have the requisite contact electrodes and software for measuring EDA.  
On the other hand, optical sensors such as user-facing cameras or photodiodes are available on almost every consumer electronics device (e.g., smartphones, smart watches, laptop computers, tablets, home Internet-of-Things devices, etc.). 
The vision of ubiquitous computing is to make computing appear anytime and everywhere and optical sensing is one way to achieve this for physiological measurement. 
Discovering novel and non-invasive methods for monitoring sympathetic arousal, beyond EDA, provides a valuable opportunity in the field of digital health, and could help people better understand stress responses.

Previous research has shown that transient responses, which correlate with a person's EDA signal, can be inferred from the periorbital, forehead, and maxillary regions of the face using thermal cameras~\cite{shastri2009edaface, pavlidis2012perinasalsweat, shastri2012perinalsastress}. 
Bhamboarae et al. provided an early proof-of-concept that a person's EDA can be inferred from the palm using an RGB camera by counting the specular reflections, which were created by specialized lighting conditions, on the palm~\cite{bhamborae2020contactlesseda}. 
Other research has shown that electrical stimulation and pain increase not only the sweat gland activity on the forehead but also increase the blood flow to the forehead (through sympathetic vasodilation) and the left cheek, and decreases the blood flow to the finger~\cite{nordin1990forehead, vassend2005electropainblood}. 
Therefore, we hypothesized that by measuring the peripheral blood flow changes to the face and hand, we should be able to infer a signal that strongly correlates with a person's tonic EDA, and that would allow us to measure a person's sympathetic arousal. 
Extensive research over the last years has shown that a person's peripheral hemodynamics, such as the blood volume pulse, blood perfusion, or vasomotion, can be inferred remotely using a regular RGB camera~\cite{wu2000photoplethysmography,takano2007heart,2011pohrppgbasics, wang2017pos, mcduff2020peripheralhemo,yu2019remote,yu2022physformer,liu2020multi, kamshilin2022bloodperfusion}, but this work has not presented comparisons with EDA. 
In addition, it is important to consider that factors such as motion~\cite{pai21hrvcam} and skin type~\cite{fallow2013influence, addison2018video, nowara2020meta} can significantly impact the performance.
Several previous works found that skin type influences the signal-to-noise ratio (SNR) of predicted rPPG signals as a larger melanin concentration absorbs more light~\cite{fallow2013influence, addison2018video, nowara2020meta} and that people with skin type \RNum{5} and \RNum{6} according to the Fitzpatrick scale~\cite{fitzpatrick1988validity} are often underrepresented in such computer vision datasets~\cite{nowara2020meta, ba2021overcoming}.

In this paper, we show that three different metrics, the total blood volume changes, the mean blood pulsation amplitude (BPA), and the instantaneous pulse rate, calculated in four different conditions each provide a measure of sympathetic arousal that is closely correlated with EDA measurements.
The four different conditions are: 1) a video of participants’ palms, 2) a video of participants’ foreheads, 3) a PPG signal from participants’ foreheads, and 4) a PPG signal from their fingers.
To this end, we contribute a self-recorded dataset in which we recorded videos of participants' faces and palms (N = 21), photoplethysmography (PPG) signals from the finger and the forehead, and EDA signals from the fingers. 
During the study, we asked the participants to alternate between relaxing and pinching their arms to cause spikes to create sympathetic responses.
In contrast to previous works, we used ambient light and a regular RGB camera instead of an expensive thermal camera. 
As input signals, we use the mean intensity pixel values from the camera feeds and the raw PPG signals from the PPG sensors.
To infer the total blood volume changes, we calculate the sum of the absolute change with a 60-second sliding window from our input signals, which we first low-pass filter to mitigate influences from, e.g., micro-expressions or respiratory changes.
To obtain the BPA, we first calculate the remote photoplethysmography (rPPG) signal from the camera feeds.
Afterward, we calculate the mean BPA from the rPPG and the contact PPG signals by again using a 60-second sliding window approach. 
We compare all three of our calculated signals with the ground truth EDA measurements recorded on the fingers and show that there is a high agreement. 
We also examine how well a person's sympathetic arousal can be inferred from different orofacial regions and how different skin tones influence the performance.

\begin{figure}[htb]
    \centering
    \includegraphics[width=1.0\textwidth]{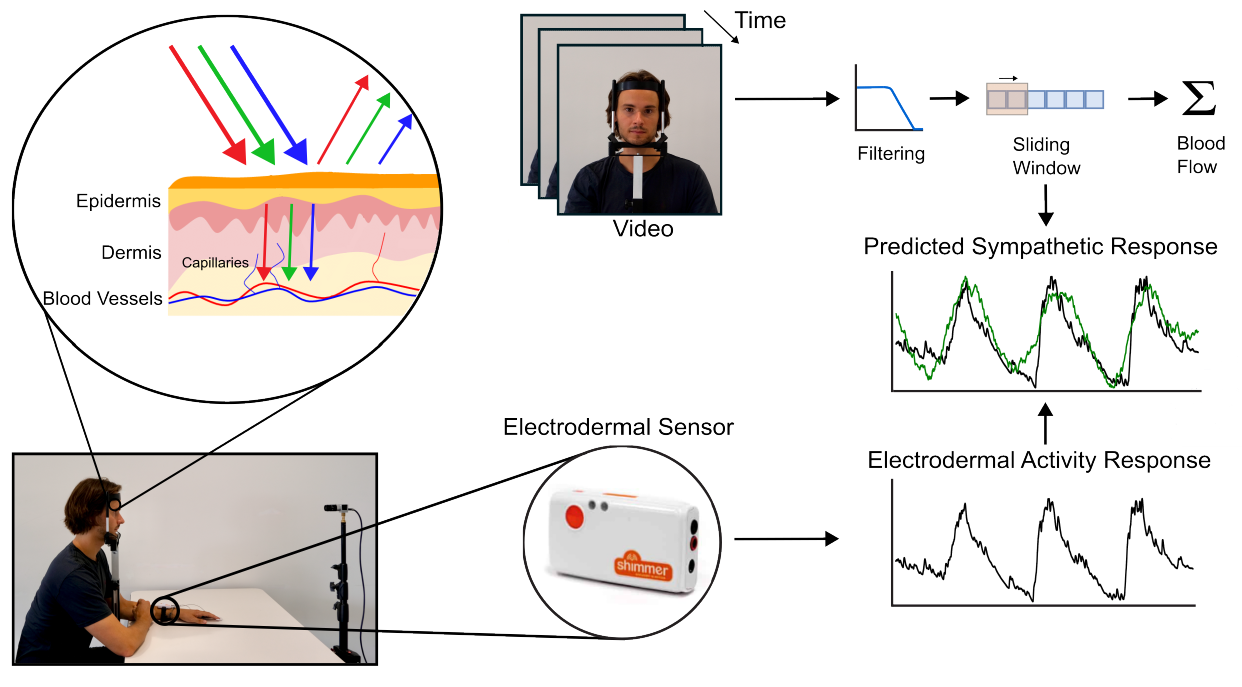}
    \caption{\textbf{A method to remotely measure a person's sympathetic arousal with a camera.} Using a sliding window approach, we estimate a person's peripheral blood flow changes from a video of the person's face or palm. We show that these blood flow changes strongly correlate with a person's electrodermal signals.}
    \label{fig:overall_figure}
\end{figure}

\section{Methods}
\subsection{Study Design}

\subsubsection{Study Protocol} 
The study protocol consisted of 3 alternating phases of 2 minutes of resting with 30 seconds of pinching and a final 2 minutes of resting (resting - pinching - resting - pinching - resting - pinching - resting). 
An experimenter instructed the participants to pinch themselves (at 2, 4.5, and 7 minutes) on the back of their arms or at their legs to stimulate a response in the participants' electrodermal activities.
With this study protocol, we ensured that the participants did not have to move their head or non-dominant hand (where the physiological signals and palm video is recorded) in any way while still causing spikes in the participants' EDA signals.
The subjects were asked not to wear any makeup and to abstain from consuming vasomotor substances (e.g., caffeine and nicotine), tranquilizers, or any psychotropic drugs for at least 4 hours prior to participating in the experiment.
The ETH Zurich Ethics Commission approved this study as proposal EK-2023-N-26.

\subsubsection{Apparatus} 
\begin{figure}[h]
    \centering
    \includegraphics[width=0.8\textwidth]{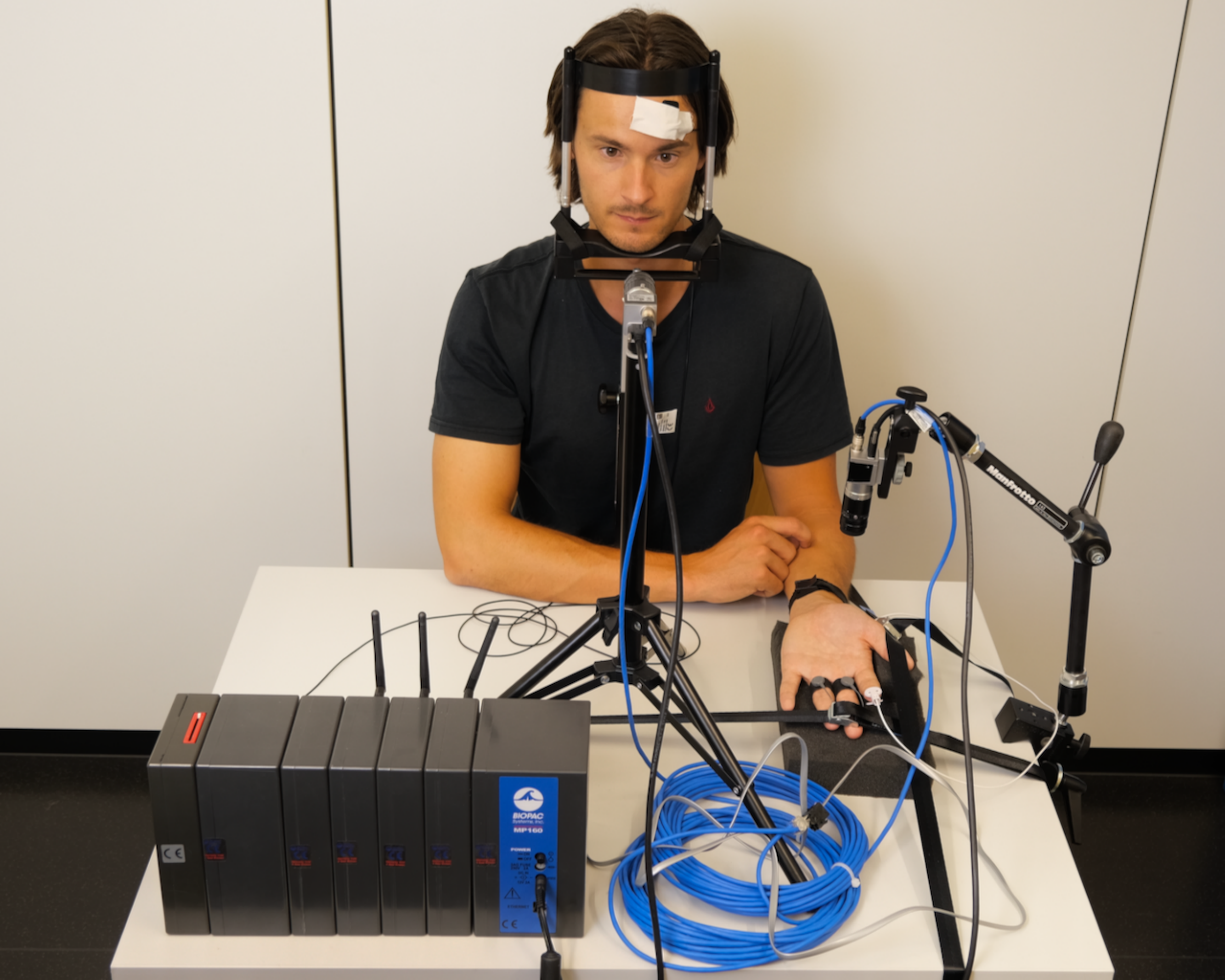}
    \caption{\textbf{The overall setup of the study.}
    The face and the palm of each participant are recorded using a Basler acA1300-200uc camera.
    The participant's EDA signal is recorded on the fingers (ring and middle finger) of the non-dominant hand using two electrodes, and the two PPG signals are recorded on the forehead and at the index finger of the non-dominant hand. 
    The participant uses the dominant hand to pinch the back of the non-dominant arm or the legs as a stressor to cause a spike in EDA.
    To minimize motion artifacts, the participant's head is placed on a chin rest, and the thumb, middle finger, and ringer are placed under a belt such that they do not interfere with the physiological measurements.}
    \label{fig:setup_overall}
\end{figure}

\begin{figure}[h]
    \centering
    \includegraphics[width=1.0\textwidth]{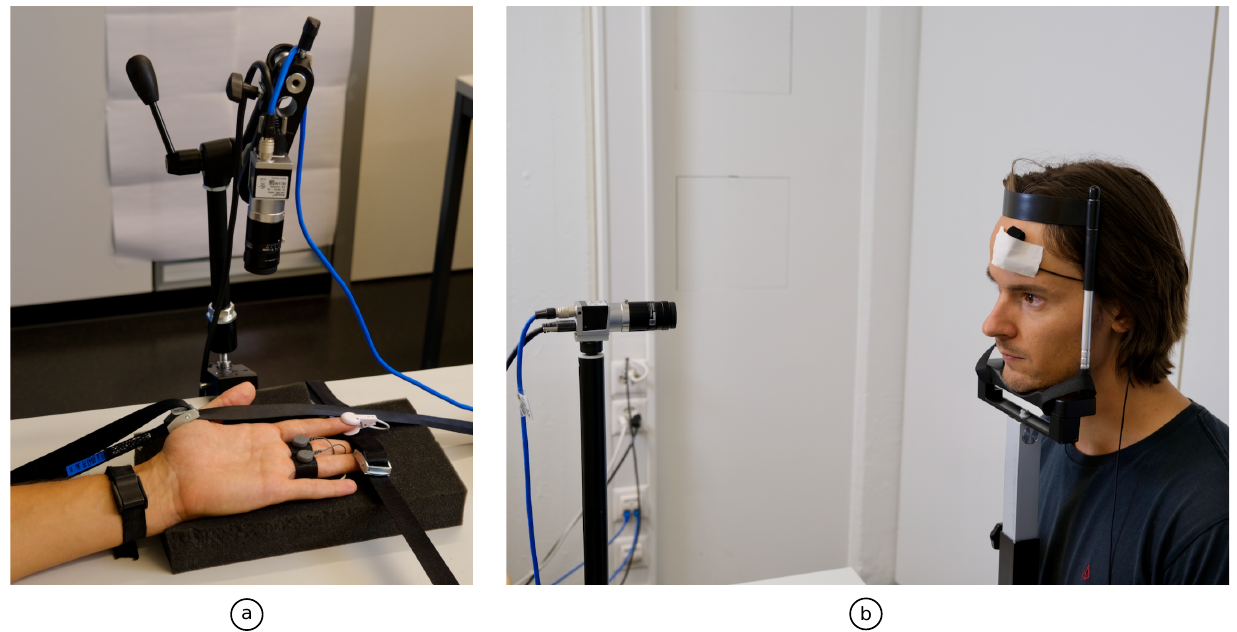}
    \caption{\textbf{The individual components of the setup.}
    (a) The camera which records the participant's palm.
    The participant's thumb, ring finger, and middle finger are placed under a belt to minimize any possible motion of the hand. 
    The participant's EDA signal and PPG signal are recorded from the index (PPG), middle (EDA), and ring fingers (EDA).
    (b) The camera which records the participant's face.
    The participant's face is placed on a chin rest to minimize any motion of the head. 
    The PPG sensor is fixated with medical tape on the forehead of the participant.}
    \label{fig:detailed_setup}
\end{figure}

We show the overall setup in \autoref{fig:setup_overall}. 
We used two Basler acA1300-200uc cameras with a frame rate of 100\,Hz to record the participants' faces and palms as we show in \autoref{fig:detailed_setup}.
The white balance and auto-focus settings were disabled throughout the experiment.
The participants' physiological signals were recorded on the non-dominant hand and the face. 
The EDA was recorded on the middle and ring fingers with two electrodes using a Shimmer GSR3+ device, and the PPG signals were recorded at the index finger (using the Shimmer GSR3+ device) and forehead (using a BIOPAC MP160 device with a BIOPAC PPG100C module).
We used the Shimmer device and the BIOPAC device to record the two PPG signals as the Shimmer GSR3+ device cannot record two PPG signals at the same time.
The Shimmer GSR3+ device was recording with a sampling rate of 100\,Hz and the BIOPAC MP160 device was recording with a sampling rate of 2000\,Hz.
We recorded the participants' PPG signals both on the finger and the forehead as previous research suggests that we should be able to see a change in blood flow both on the finger and the face~\cite{nordin1990forehead, vassend2005electropainblood}. 
To minimize any motion artifacts, an experimenter instructed the participants to place their heads on a chin rest and to place their thumbs, ring fingers, and middle fingers under a belt as we show in \autoref{fig:setup_overall} and \autoref{fig:detailed_setup}.
To synchronize the camera recordings with the physiological signals recorded by the Shimmer device and the BIOPAC, we analogy triggered the cameras from the BIOPAC device with a frequency of 100\,Hz. 
In this way, the PPG measurements from the BIOPAC device and the camera recordings are synchronized. 
To synchronize the BIOPAC measurements (and therefore the camera measurements) with the Shimmer recordings, we align the signals in a post-processing step using Unix timestamps recorded by the BIOPAC and the Shimmer device.

\subsubsection{Population} 

For our study, we recorded nine and a half minutes of measurements from 21 individuals (5 female, 16 male) between the ages of 19 and 36. 
The mean age of the participants was 26.4 years, with a standard deviation of 3.9 years.
According to the Fitzpatrick scale~\cite{fitzpatrick1988validity}, 5 individuals have skin type \RNum{2}, 11 individuals skin type \RNum{3}, 3 individuals skin type \RNum{5}, and 2 individuals skin type \RNum{6}.

\subsubsection{Environmental Conditions}

\begin{figure}[h]
    \centering
    \includegraphics[width=1\textwidth]{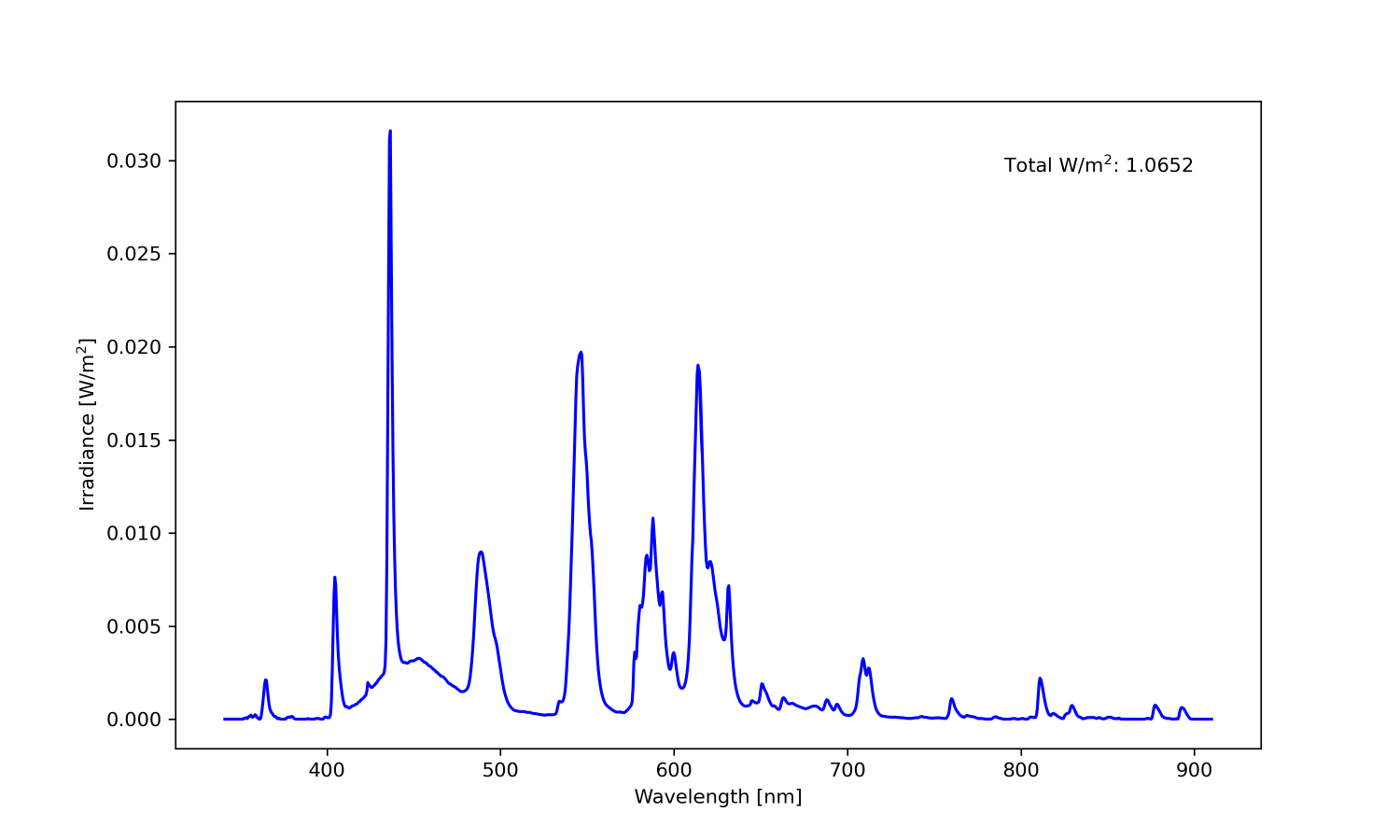}
    \caption{\textbf{The spectral power distribution of the lighting in the study room.}
    The lighting conditions were the same for all study participants.
    We always used the same lighting source and a room without windows.
    The spectral power distribution was measured using a spectroradiometer (StellarNet BLUE-Wave).}
    \label{fig:radiometer}
\end{figure}

The study was conducted in an air-conditioned room with a constant temperature between 24 and 25\,°C.
For all participants, the lighting conditions were kept the same as we always used the same lighting source (a standard ceiling light) and a room without windows.
To make the results reproducible, we measured the spectral power distribution (see ~\autoref{fig:radiometer}) in the study room using a spectroradiometer (StellarNet BLUE-Wave).

\subsection{Preprocessing}

\subsubsection{Ground Truth EDA Signal}
To keep the EDA signal as unprocessed as possible, we only apply a 2\textsuperscript{nd} order high pass Butterworth filter to the ground truth EDA signal with a cutoff frequency of 0.003\,Hz and then normalize the signal between 0 and 1 to compare it with the inferred sympathetic responses.
In this way, we can filter out the very slow increase of the EDA signal caused e.g., by sweating due to the electrodes on the finger, which can be seen in the top plot of \autoref{fig:preprocessing_eda} (b).
To determine which participants have a significant change in their EDA signal, we use a dependent t-test for paired samples to check if the mean EDA values before the pinching are significantly different than the mean EDA values during the 30 seconds of pinching. 
Of the 120-second periods before each pinching, we only use the last 90 seconds to consider that the increase in EDA can be delayed by up to 20 seconds after the pinching. 
For 17 out of the 21 participants, the p-value is smaller than 0.05. 
For participants 01, 02, 12, and 14, the p-value is bigger than 0.05. Therefore, we excluded them from our quantitative analysis.
As the dependent t-test for paired samples requires normality, we used the Shapiro-Wilk test to ensure that the calculated mean values follow a normal distribution. We obtain a p-value of 0.14 which means that our mean values follow a normal distribution as the null hypothesis that the data was drawn from a normal distribution cannot be rejected (p-value > 0.05).

\subsubsection{Input Signals}
\label{sec:input_signals}
\begin{figure}[h]
    \centering
    \includegraphics[width=0.5\textwidth]{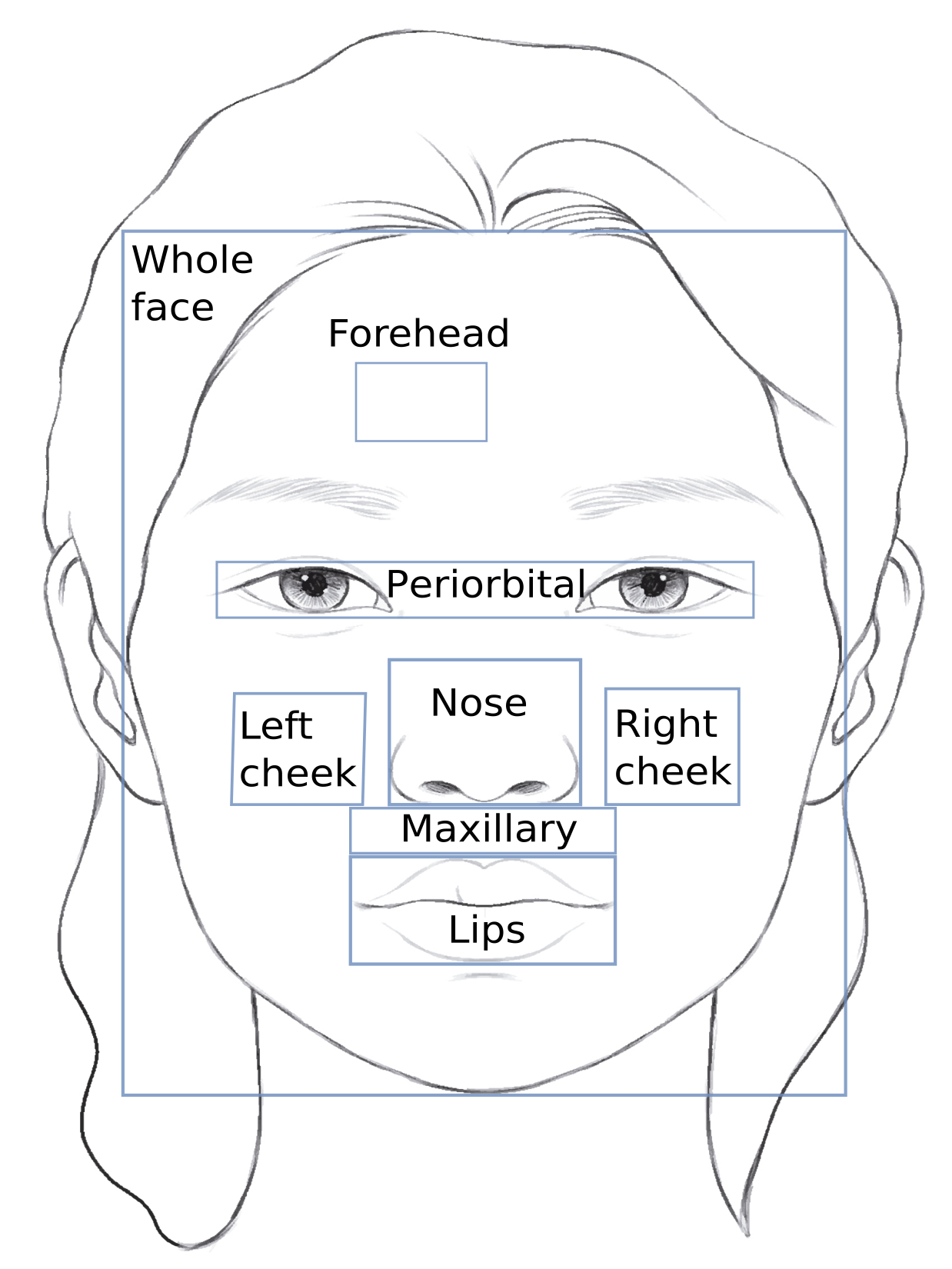}
    \caption{\textbf{The bounding boxes of the orofacial regions.}
    This figure shows where we placed the different bounding boxes on the face to analyze how well a person's sympathetic response can be measured from the different orofacial regions on the face.}
    \label{fig:facial_regions}
\end{figure}

As input signals, we have the videos of the participants' faces and palms and the recordings of the PPG sensors on the participants' foreheads and fingers.
We used two different devices to record the two PPG signals as the Shimmer GSR3+ device, which is used to record the EDA and PPG on the finger, cannot record two PPG signals at the same time.
The videos and the PPG signal from the finger (using the Shimmer device) are recorded with a frequency of 100\,Hz, while the PPG signal from the forehead (using the BIOPAC device) is recorded with a frequency of 2000\,Hz.
Therefore, we downsample the PPG signal from the forehead to 100\,Hz to synchronize it with the PPG signal from the finger.
From the videos, we calculate the mean values for the red, green, and blue channels of the camera for every frame, and for the PPG sensors, we use the recorded raw signal. 
As we expect the signal strength of both the hemodynamics and the rPPG signals to vary spatially, we evaluate different orofacial regions from the videos of the participants' faces to analyze from which orofacial region the sympathetic arousal can be inferred best.
Due to individual variations in facial shapes and sizes, we carefully select each participant's cropping regions by hand to maintain objectivity and consistency.
As the participants could not move their heads while their heads were on the chin rest during the study, we did not have to use any facial tracking and were able to select the cropping regions by hand.
We show the different orofacial regions in \autoref{fig:facial_regions}, which are the whole face, forehead, right cheek, left cheek, nose, lips, maxillary, and periorbital region.
Before manually cropping the videos to the different orofacial regions, we resize all videos to a size of 256 x 205 (width x height) pixels. 
In \autoref{tab:orofacial_regions_positioning}, we explain in detail how we individually positioned each of the bounding boxes for the different orofacial regions when frontally looking at the person’s face.

\begin{table}[h]
    \centering
    \renewcommand{\arraystretch}{1}
    \begin{tabular}{ll}
    \toprule
         \textbf{Region} & \textbf{Positioning of the bounding boxes} \\
         \midrule 
         Whole face & - Left and right: Widest points of the face, not including the ears \\
         & - Bottom and top: Between the chin and start of the hair line \\
         \midrule 
         Forehead & - Bounding box of size (30 x 15) pixels \\
         & - Vertically centered at the midpoint between the start of the hair line and the \\
         &   midpoint between the eyebrows \\
         & - Right edge of the bounding box along the midpoint between the eyebrows \\
         \midrule 
         Right cheek & - Left and right: 5 pixels to the right from the end of the right nostril with a \\
         & width of 25 pixels \\
         & - Bottom and top: Between the bottom of the nose until the mid between the \\ 
         & bottom of the nose and the midpoint of the eyes \\
         \midrule 
         Left cheek & - Left and right: 5 pixels to the left from the end of the left nostril with a \\
         & width of 25 pixels \\
         & - Bottom and top: Between the bottom of the nose until the mid between the \\ 
         & bottom of the nose and the midpoint of the eyes \\
         \midrule 
         Nose & - Left and right: Widest points of the left and right nostrils \\
         & - Bottom and top: Between the bottom of the nose until two thirds of the \\
         & way to the midpoint between the eyes \\
         \midrule 
         Lips & - Left and right: End points of the lips \\
         & - Bottom and top: Widest points of the lower and upper lip \\
         \midrule 
         Maxillary & - Left and right: End points of the lips \\
         & - Bottom and top: Highest point of the upper lip and bottom of the nose \\
         \midrule 
         Periorbital & - Left and right: Most left point of the left and most right point of the right eye \\
         & - Bottom and top: Lowest point of both eyes and highest point of both eyes \\
         \bottomrule
    \end{tabular}
    \caption{\textbf{Positioning of the bounding boxes.}
    This table explains how we individually positioned each of the bounding boxes for the different orofacial regions. The descriptions are written from the perspective of frontally looking at the person's face as displayed in \autoref{fig:facial_regions} and bounding box sizes are given in pixels (width x height).}
    \label{tab:orofacial_regions_positioning}
\end{table}

\subsection{Sympathetic Arousal Assessment}
To assess the participants' sympathetic arousal, we compare three different methods, the \textit{total blood volume change}, the \textit{mean blood pulsation amplitude} and the \textit{instantaneous pulse rate}.
We apply all three methods to all four input modalities: 
1) the video of participants’ palms, 2) the video of participants’ foreheads, 3) the PPG signal from participants’ foreheads, and 4) the PPG signal from the participants' fingers.
The \textit{total blood volume change} method uses as input the raw mean intensity pixel values from the camera feeds and the raw contact PPG signals, which are both only low-pass filtered to mitigate influences, e.g., from micro-expression, a person's heart rate, or a person's respiration.
The \textit{mean blood pulsation amplitude} method and the \textit{instantaneous pulse rate} method both use a person's PPG signal as an input.
Therefore, we first calculate the remote photoplethysmography (rPPG) signal from the camera feeds by using an AC/DC filtering approach for the \textit{mean blood pulsation amplitude} and a Butterworth bandpass filtering approach for the \textit{instantaneous pulse rate}.
We apply the same filters to the contact PPG signals to obtain clean PPG signals.

\subsubsection{Total Blood Volume Change}
\label{sec:absolute_change_method}
Previous research has shown that as a response to pain, the blood flow to the forehead and the left cheek increases, the blood flow to the finger decreases, and the electrodermal activity increases~\cite{nordin1990forehead, vassend2005electropainblood}.
Given that variations in blood flow volume influence the measurements from cameras~\cite{2010pohrppgbasics, 2011pohrppgbasics} and contact PPG sensors~\cite{huiwen2022bloodflow}, we expect that both of these measurements have an increase in the total change during periods of pain compared to periods of relaxation.
For individuals with an increase in EDA during periods of pain, the EDA signal should therefore correlate with the total change of the measurements from the cameras and contact PPG sensors.

The proposed method calculates the absolute change of the input signal $s(n)$, which can either be the intensity pixel values from the camera or the measurements from the contact PPG sensors. 
By calculating the absolute change, this method is applicable to regions where blood flow either increases or decreases.
In our case, $n$ is defined as $n \in [0, N]$ with $N=57000$ as all of our input signals have a recording length of 9 and a half minutes with a sampling rate of 100\,Hz.
Before calculating the absolute change, we first filter the input signal $s(n)$ with a 2\textsuperscript{nd} order Butterworth low pass filter with a cutoff frequency of 0.05\,Hz.
In this way, we ensure to minimize all influences to our signal from motion such as macro-expressions (duration between 0.5 and 4.0\,seconds~\cite{ekman2004emotions, matsumoto2011evidence}) or micro-expressions (duration between 0.067 and 0.5\,seconds~\cite{yan2013microexpressions, matsumoto2011evidence}), a person's heart rate (frequencies higher than 0.7\,Hz corresponding to 42\,bpm~\cite{spodick1992operational, heartrateaha}), a person's normal respiratory rate (0.2 to 0.33\,Hz~\cite{flenady2017respiratoryrate}), and non-linear dynamics of human respiration or Mayer waves (frequencies of 0.1\,Hz or higher~\cite{wysocki2006nonlinearbreathing, claude2006mayerwaves}).
The remaining signal should then be dominantly influenced by a person's blood flow or sweat changes.
To calculate the absolute change of the input signal $s(n)$, we use a sliding window approach with a window size of $w=6000$ (corresponding to 60 seconds), with a stride of 1 sample (corresponding to 0.01\,Hz).
Therefore, we obtain $K = N - w + 1 = 51001$ sliding windows for each input signal.
For each sliding window $SW_{k}$, with $k \in [0, K]$, we then calculate the sum of the absolute change of the input signal $s(n)$ as:
\begin{equation}
SW_{k} = \sum_{n=k}^{w+k} \left|s(n+1)-s(n)\right|\
\end{equation}
Finally, we filter the resulting signal with the same 2\textsuperscript{nd} order Butterworth high pass filter that we used for the ground truth EDA signal with a cutoff frequency of 0.003\,Hz and normalize the inferred signal between 0 and 1 to compare it with the ground truth EDA signal. 
As the sliding window approach shortens the obtained signal by $N-K=5999$ samples, we cut off the first $2999$ samples (corresponding to 29\,seconds) and the last $3000$ samples (corresponding to 30~seconds) of the ground truth EDA signal.


\subsubsection{Mean Blood Pulsation Amplitude}
\label{sec:bloodpulsation}
Kamshilin et al.~\cite{kamshilin2022bloodperfusion} have shown that the blood perfusion of tissue can be inferred by calculating the mean blood pulsation amplitude from the rPPG signal from a camera.
We adapt this approach to our setup as we assume that the blood flow to the face and the finger changes as a response to pain (see Section~\ref{sec:absolute_change_method}).
In a first step, we calculate the rPPG signal from the camera (see Equation~\ref{eq:1}). 
We emphasize the alternating component (AC) of the rPPG signal by dividing the input signal $s(n)$ from the camera feed by the slowly varying DC component of the rPPG signal.
As explained above, $n$ is defined as $n \in [0, N]$ with $N=57000$.
We calculate the DC component of the rPPG signal by calculating the mean of the intensity pixel values for each recording.
Then, we subtract the unity from the AC/DC ration and invert the sign to obtain a signal that positively correlates with the blood perfusion in the face.
To remove high-frequency noise in the signal, we use a Butterworth high-pass filter with a cut-off frequency at 4\,Hz.
In a second step, we calculate the amplitude of the rPPG signal by using a sliding window approach with a window size of 1 second.
For each window, we take the maximum and minimum value of the input signal to obtain the pulse amplitude signal.
In a third step, we take the mean of the pulse amplitude signal with a sliding window approach of 60 seconds to obtain the mean blood pulsation amplitude. 
Our obtained signal has, as explained above, a length of $K = N - w + 1 = 51001$ sliding windows.
Kamshilin et al.~\cite{kamshilin2022bloodperfusion} used a dynamic sliding window approach that was based on the R-peaks of the recorded ECG signal and not a static window size of 60 seconds.
However, as we did not record the participants' ECG signal in our study, we used a static window size of 60 seconds.

\begin{equation}
\label{eq:1}
rPPG(n) = - \left (\frac{AC}{DC} -1 \right) = - \left(\frac{s(n)}{\frac{1}{n}\sum_{n=0}^{N} s(n)}-1 \right)
\end{equation}

\subsubsection{Instantaneous Pulse Rate}
\label{sec:inst}
Kettunen et al.~\cite{keettunen1998edahrcorrelation} have shown that a person's EDA and instantaneous pulse rate (calculated using the duration of the interbeat intervals) are synchronized.
This method, therefore, calculates a person's instantaneous pulse rate from the input videos (by calculating the rPPG signal) or PPG signals to infer a person's EDA.
First, we filter the input signal $s(n)$ with a 2\textsuperscript{nd} order Butterworth band pass filter with cutoff frequencies at 0.7\,Hz and 2.5\,Hz (corresponding to [42, 150]\,bpm)~\cite{liu2020mttscan}.
As explained above, $n$ is defined as $n \in [0, N]$ with $N=57000$.
To decrease the influence of motion and improve the performance for videos with dark skin, we use the approach from Pai et al.~\cite{pai21hrvcam} to reconstruct a cosine signal that estimates the instantaneous frequency of the rPPG signal before calculating the interbeat intervals.
Afterward, we determine the pulse rate peaks to calculate the interbeat intervals.
Then, we use a sliding window approach with a window size of $w=60$ interbeat intervals (corresponding to 0.4 to 1.4\,minutes depending on the pulse rate) with a stride of 1 interbeat interval to calculate the mean duration of the interbeat intervals $IBI$ of the corresponding sliding window. 
Depending on the person's pulse rate, we obtain $M$ interbeat intervals resulting in $K = M - w + 1$ sliding windows.
For each sliding window $SW_{k}$, with $k \in [0, K]$, we then calculate the mean duration of the interbeat intervals as:
\begin{equation}
SW_{k} = \frac{1}{M} \sum_{n=1}^{M} IBI(n)\
\end{equation}
Subsequently, we filter the resulting signal with the same 2\textsuperscript{nd} order Butterworth high pass filter that we used for the ground truth EDA signal with a cutoff frequency of 0.003\,Hz and normalize the inferred signal between 0 and 1 to be able compare it with the ground truth EDA signal. 
Finally, as we expect that the mean duration of the IBIs decreases when the EDA increases~\cite{keettunen1998edahrcorrelation}, we flip the signal by subtracting it from one.

\subsection{Motion Analysis}
\label{sec:motion}
During the phases of high arousal caused by self-pinching, it is possible that not only the physiological signals, such as the blood flow or EDA, change but also that the participants have increased facial motion due to the pain.
To, therefore, assess whether the predicted signals from our three proposed approaches are affected by motion, we calculate the magnitude of the optical flow~\cite{horn1981determining} for all orofacial regions and compare them to our predicted signals using the Spearman correlation.
As some of our orofacial regions, such as the forehead or cheeks, do not have distinctive corner points, we use the Farneback method to compute the dense optical flow for all pixels in each image~\cite{farneback2003two}.
To capture fast motions such as micro-expressions (duration between 0.067 and 0.5\,seconds~\cite{yan2013microexpressions, matsumoto2011evidence}), we then calculate the mean optical flow with the same sliding window approach that we used in our three proposed approaches (window length of 60 seconds, stride of 1 frame).

\subsection{Evaluation Metrics}
\label{sec:evaluation_method}

We use two different evaluation metrics to compare our predicted signals with the ground truth EDA signals: the Spearman correlation and the weighted Kendall's $\tau$ correlation.
To compare our inferred signals with the ground truth EDA signals, we use the Spearman correlation coefficient as it does not require that our data be distributed normally like, e.g., the Pearson correlation coefficient requires.
As our sliding window approach can cause a time shift in our inferred EDA signal, we slide the ground truth EDA signal for a maximum of 20 seconds according to the maximum cross correlation between the ground truth EDA signal and our inferred EDA signal.
Furthermore, to assess whether our approach is capable of capturing different levels of sympathetic arousal, which is the ultimate goal of our approach, we use the weighted Kendall's $\tau$ correlation with the hyperbolic weighing function.
We divide the ground truth EDA signals and our predicted signals into three ranks (low, medium, and high arousal) and then compute the weighted Kendall's $\tau$ correlation.

\section{Results}
\subsection{Validation of Sympathetic Arousal Induction}
\label{sec:effect_pinching}

\begin{figure}[H]
    \centering
    \includegraphics[width=1\textwidth]{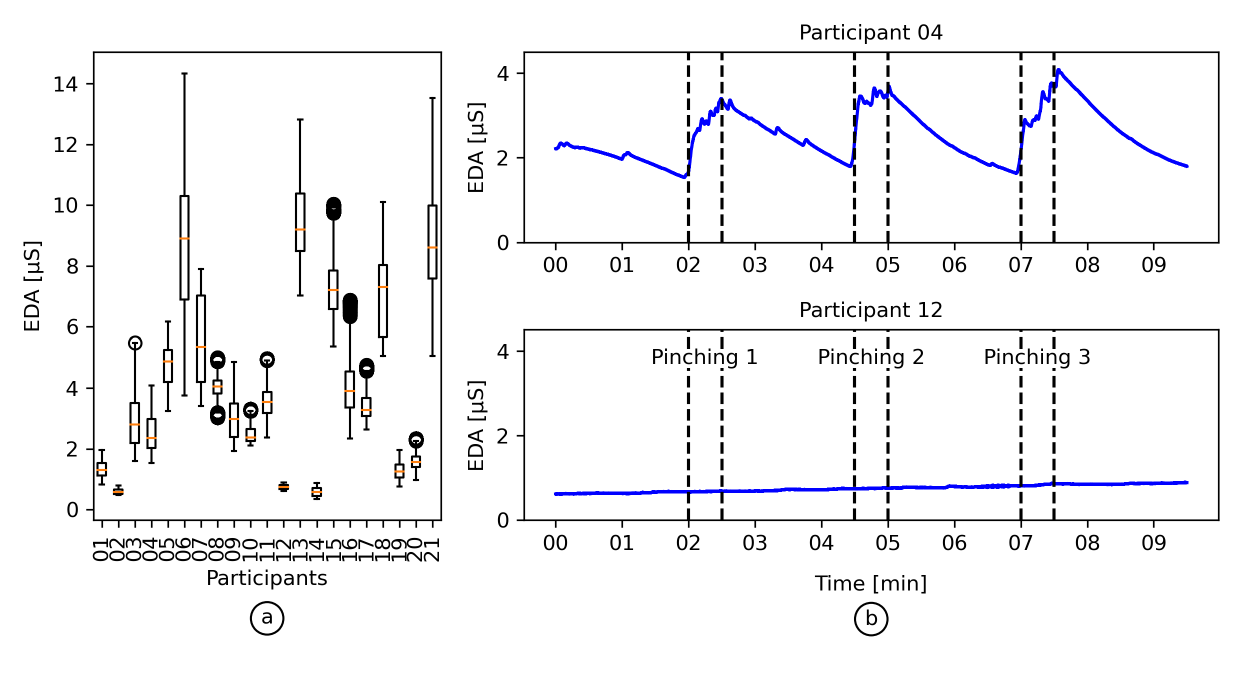}
    \caption{\textbf{Distributions of the participants' EDA levels and two example responses to the stressors.}
    (a) shows the distribution of the absolute EDA values of all 21 participants of the study.
    Three of the four participants, 02, 12, and 14, which did not have a significant change in their EDA values before and during the stressors, also have the lowest EDA values.
    (b) shows two example contact EDA signals from participants 04 and 12 for the entire duration of the study. 
    In the top plot, the spikes in EDA during the stressor periods are clearly visible for participant 04.
    In the bottom plot, no such changes in EDA are visible for participant 12.}
    \label{fig:preprocessing_eda}
\end{figure}

The mean tonic EDA level across participants was \SI{3.93}{\micro\siemens} and the variance was \SI{2.90}{\micro\siemens} (see box plots in \autoref{fig:preprocessing_eda}a).  
Four (19\%) of the 21 study participants did not have a significant response in their EDA before or during the stressor (dependent t-test for paired samples, \textit{p} > 0.05). 
Three of these four participants also had the lowest tonic EDA values (see \autoref{fig:preprocessing_eda}a).
We expected such differences in the participants' EDA level as prior work has shown that EDA has a high inter-subject variability~\cite{posadoquintero2016edafrequency}.
Examples of the contact EDA responses (\autoref{fig:preprocessing_eda}b) illustrate these patterns. 
Sympathetic arousal during the three stressor periods is clearly visible for one participant and no such changes in the EDA signal are visible for another.
For clarity, we split the analysis of these four participants who exhibited no arousal responses from those who did in our subsequent results.
In previous research, similar results were observed where three out of 14 participants did not have changes in their EDA on the forehead as a response to electrical stimulation of the median nerve at the wrist~\cite{nordin1990forehead}.

\subsection{Inferring EDA from Blood Flow Changes and Instantaneous Pulse Rate}
\label{sec:inferringEDA}

In the following analysis, we evaluate how much of the variance within the electrodermal responses can be explained by videos recorded from the palm and face or by contact PPG signals measured from the finger and forehead.
To do so, we quantify the correlation between the ground truth EDA signals and the signals obtained from our three methods, the total blood volume changes, the mean blood pulsation amplitude and the instantaneous pulse rate.
All three signals, the total blood volume change, the mean blood pulsation amplitude, and the instantaneous pulse rate, can be inferred using either only the recorded videos or using only the measurements from the contact PPG sensors (see Methods section).
As our goal is to test a non-contact approach, we put emphasis in our analysis on the camera measurements and validate the results from the camera measurements with the results from the contact PPG sensors. This removes doubt that the camera is measuring signals that a contact PPG cannot.

We show the calculated total blood volume changes, the mean blood pulsation amplitude, and the instantaneous pulse rates compared to their ground truth EDA signals for four participants and four different input signals in \autoref{fig:example_plots_good}.
The correlations between our calculated, displayed signals and the ground truth signals are between 0.50 and 0.75.
We found that the calculated signals from all three of our methods closely follow the tonic signal of the ground truth EDA signal.
However, none of the three proposed methods do capture smaller phasic changes in the ground truth EDA signal.

\subsubsection{Camera Color Channels}
For the analysis of the recorded camera feeds, we evaluate the red, green, or blue channels separately to identify which color channel contains the strongest signal.

\textbf{Total blood volume change.}
Absolute changes in total blood volume had the highest median correlation from the green channel (median correlation $\rho$ across all participants was maximum for forehead region with $\rho = 0.63$, all correlations $p < 0.01$ except for participant 10) followed by the blue channel (median correlation $\rho$ across all participants was maximum for forehead region with $\rho = 0.40$, all correlations $\textit{p} < 0.01$) and then the red channel (median correlation $\rho$ across all participants was maximum for palm region with $\rho = 0.31$, all correlations $\textit{p} < 0.01$).
Therefore, we will use the green channel for calculating the total blood volume changes.

\textbf{Mean blood pulsation amplitude.}
The highest median correlation from the mean blood pulsation amplitude was from the red channel (median correlation $\rho$ across all participants was maximum for forehead region with $\rho = 0.58$, all correlations $p < 0.01$) followed by the the green channel (median correlation $\rho$ across all participants was maximum for forehead region with $\rho = 0.58$, all correlations $\textit{p} < 0.01$) and then the blue channel (median correlation $\rho$ across all participants was maximum for periorbital region with $\rho = 0.46$, all correlations $\textit{p} < 0.01$).
Therefore, we will use the red channel for calculating the mean blood pulsation amplitude.

\textbf{Instantaneous pulse rate.}
Instantaneous pulse rate also had the highest median correlation from the green channel (median correlation across all participants was maximum for nose region with $\rho = 0.49$, all correlations $\textit{p} < 0.01$), followed by the red channel (median correlation across all participants was maximum for nose region with $\rho = 0.46$, all correlations $\textit{p} < 0.01$) and then the blue channel (median correlation across all participants was maximum for right cheek region with $\rho = 0.12$, all correlations $\textit{p} < 0.01$).
Therefore, we will use the green channel for calculating the instantaneous pulse rate in our following analysis. These results are supported by prior works that show that the green channel has the highest signal-to-noise ratio for measuring a person's rPPG signal under normal lighting conditions (daylight and artificial indoor light [e.g., incandescent bulbs])~\cite{verkruysse2008ppgcolor}.

\subsubsection{Regions of Interest}
\label{sec:region_interest}
We expect the signal strength of both the hemodynamics and the rPPG signals to vary spatially in the video of the face.
Therefore, we separate the face into different orofacial regions as we show in \autoref{fig:facial_regions}.
Prior work has shown that as a response to pain, the blood flow to the forehead and the left cheek increases, and the blood flow to the finger decreases~\cite{nordin1990forehead, vassend2005electropainblood}.

\begin{figure}[H]
    \centering
    \includegraphics[width=1.0\textwidth]{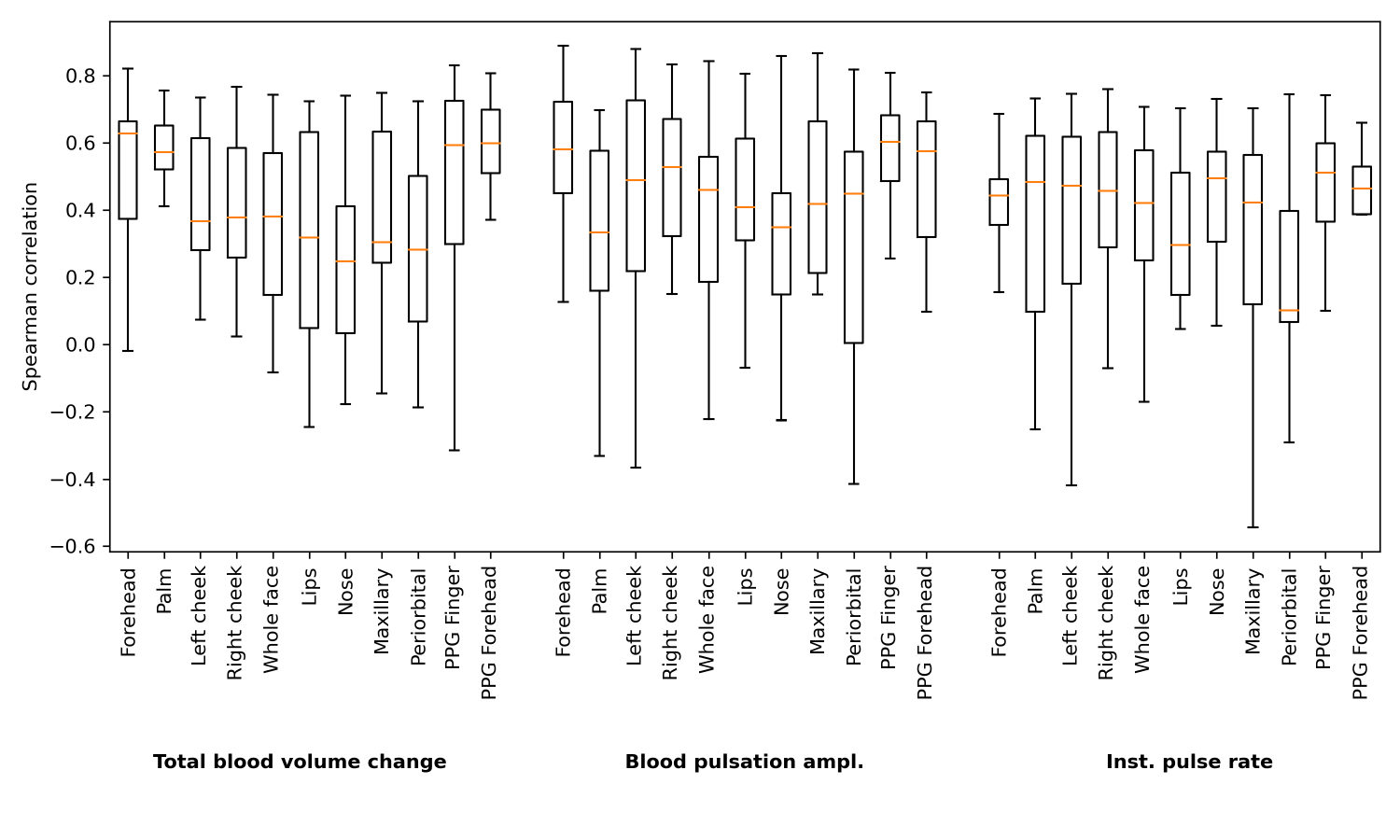}
    \caption{\textbf{Box plot of the Spearman correlations between the calculated signals and the ground truth EDA signals.}
    This shows the Spearman correlations between the calculated signals and the ground truth EDA signals across all participants and for all regions using the total blood volume change, the mean blood pulsation amplitude, and the instantaneous pulse rate.
    The regions are ordered in descending order for the median correlations obtained using the total blood volume changes with the results from the two PPG sensors as last for all three methods.}
    \label{fig:boxplot}
\end{figure}

\textbf{Total blood volume change.}
For the camera feeds, the median correlations across all participants between the ground truth EDA and the total blood volume changes were highest for the videos of the foreheads (median correlation $\rho = 0.63$, all correlations $p < 0.01$ except for participant 10) and palms ($\rho = 0.57$, all correlations $\textit{p} < 0.01$).
From the videos of the face, the maximum individual correlation was from the foreheads with a correlation of 0.82, $\textit{p} < 0.01$ and for the videos of the palms the maximum individual correlation was 0.72, $\textit{p} < 0.01$.
For the contact PPG sensors, the median correlation across all participants was also highest on the forehead ($\rho = 0.60$, all correlations $\textit{p} < 0.01$), followed by the finger ($\rho = 0.59$, all correlations $\textit{p} < 0.01$) (see \autoref{tab:RGB_overall_results}).
Using the contact PPG sensors, the maximum individual correlation from the finger was 0.83, $\textit{p} < 0.01$ and the maximum individual correlation from the forehead was 0.81, $\textit{p} < 0.01$.
From the maxillary, periorbital, and nose regions, we obtained considerably lower median correlations across all participants between $\rho = 0.25$ and $\rho = 0.30$.
The variance across all participants is smallest for the video of the palm and the PPG sensor on the forehead (see \autoref{fig:boxplot} for the distribution of the data).

\textbf{Mean blood pulsation amplitude.}
We also obtain the strongest correlation from the videos of the forehead (median correlation across all participants $\rho = 0.58$, all correlations $p < 0.01$) followed by the right cheek (median correlation across all participants $\rho = 0.53$, all correlations $p < 0.01$).
The maximum individual correlation was from the video of the forehead with a correlation of 0.89, $\textit{p} < 0.01$ followed by a video of the left cheek with a correlation of 0.88, $\textit{p} < 0.01$.
For the contact PPG sensors, the median correlations was highest from the PPG sensor on the finger ($\rho = 0.62$, all correlations $p < 0.01$).
The maximum individual correlation was 0.81, $p < 0.01$ from the contact PPG sensor on the finger.
Compared to the total blood volume change, the median and mean correlations are more similar across all regions.

\textbf{Instantaneous pulse rate.}
We expect the strongest rPPG signals in the forehead and cheeks to calculate the instantaneous pulse rate~\cite{kwon2015rppgfacialregions, kim2021rppgfacialregions}. 
For the recorded videos, the highest median correlation was obtained from the video of the nose ($\rho = 0.49$, all correlations $\textit{p} < 0.01$), closely followed by the videos of the left cheek ($\rho = 0.47$, all correlations $\textit{p} < 0.01$) and the right cheek ($\rho = 0.46$, all correlations $\textit{p} < 0.01$).
The maximum individual correlation was from the video of the right cheek and the forehead with a correlation of 0.76, $\textit{p} < 0.01$.
The median correlation from the contact PPG sensors was highest from the finger ($\rho = 0.53$, all correlations $\textit{p} < 0.01$) with a maximum individual correlation of 0.74, $\textit{p} < 0.01$.
The lowest variance is from the video of the forehead and the PPG sensor on the forehead and the finger (see \autoref{fig:boxplot}).

\begin{table}[H]
    \centering
    \begin{tabular}{llcccccc}
    \toprule
         \textbf{Sensor} & \textbf{Region} & \multicolumn{2}{c}{\textbf{Blood vol. change}} & \multicolumn{2}{c}{\textbf{Blood puls. ampl.}} & \multicolumn{2}{c}{\textbf{Inst. pulse rate}}\\
         & & Median & Mean & Median & Mean & Median & Mean\\
         \midrule 
         \multirow{9}{*}{Camera} & Forehead & \textbf{0.63} & 0.49 & \textbf{0.58} & \textbf{0.59} & 0.44 & 0.40\\
         & Palm & 0.57 & \textbf{0.55} & 0.45 & 0.38 & 0.33 & 0.32\\
         & Right cheek & 0.38 & 0.41 & 0.53 & 0.49 & 0.46 & \textbf{0.40} \\
         & Whole face & 0.38 & 0.36 & 0.46 & 0.39 & 0.42 & 0.38\\
         & Left cheek & 0.37 & 0.43 & 0.49 & 0.40 & 0.47 & 0.35\\
         & Lips & 0.32 & 0.30 & 0.41 & 0.42 & 0.30 & 0.30\\
         & Maxillary & 0.30 & 0.37 & 0.42 & 0.47 & 0.42 & 0.29\\
         & Nose & 0.25 & 0.24 & 0.35 & 0.29 & \textbf{0.49} & 0.42\\
         & Periorbital & 0.28 & 0.30 & 0.45 & 0.32 & 0.10 & 0.21\\
         &&&&& \\
         \multirow{2}{*}{PPG sensor} & Forehead & \textbf{0.60} & \textbf{0.55} & 0.58 & 0.47 & 0.46 & 0.40\\
         & Finger & 0.59 & 0.49 & \textbf{0.62} & \textbf{0.60} & \textbf{0.47} & \textbf{0.43} \\
         \bottomrule
    \end{tabular}
    \caption{\textbf{Median and mean Spearman correlations between the ground truth EDA signals and the total blood volume change, the mean blood pulsation amplitude, and the instantaneous pulse rate.}
    This table shows the median and mean Spearman correlations across all participants between the contact EDA and the calculated signals from all regions and sensors using the total blood volume change, the mean blood pulsation amplitude, and the instantaneous pulse rate method.
    The results are ordered in descending order for the median correlations obtained using the total blood volume changes.}
    \label{tab:RGB_overall_results}
\end{table}

\begin{figure}[H]
    \centering
    \includegraphics[width=0.8\textwidth]{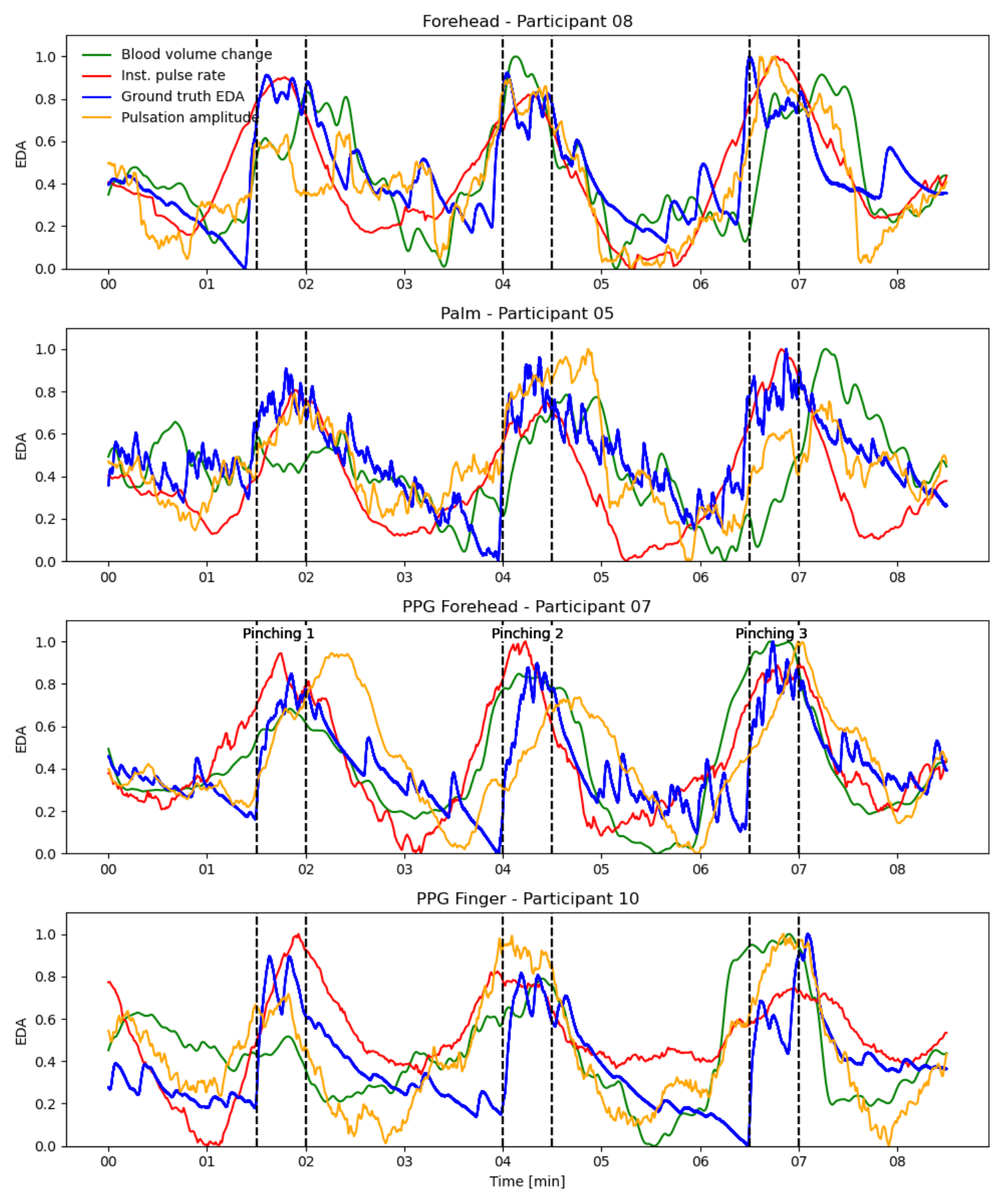}
    \caption{\textbf{Visual comparison between the ground truth EDA and the calculated sympathetic responses.}
    This plot shows the ground truth EDA signals, the calculated total blood volume changes, and the calculated instantaneous pulse rates for four exemplary participants using the four different input signals for four representative participants.
    The correlations between the calculated signals and the ground truth EDA signals are between 0.50 and 0.75.}
    \label{fig:example_plots_good}
\end{figure}

\subsubsection{Robustness to Skin Tone}
\label{sec:robustness_skin_tone}
We also analyze the performance differences for different skin tones by comparing the obtained performance when using all participants of our study ($N=17$) compared to only using participants with, according to the Fitzpatrick scale~\cite{fitzpatrick1988validity}, individuals with type \RNum{2} and \RNum{3} ($N=12$).
Using the total blood volume change method, we generally see a decrease in performance.
While the median correlation stays the same for the video of the forehead, the mean correlation decreases by 4\%.
For the PPG sensors on the forehead/ finger, the median correlations decrease by 2\%/ 8\% and the mean correlations decrease by 3.5\%/ 12\%.
The instantaneous pulse rate is even more influenced with a decrease in median correlation of 6\% for the videos of the forehead and a decrease of 4\%/ 11\% for the PPG sensors on the forehead/ finger.
On the other hand, using the mean blood pulsation amplitude on the videos of the forehead, the median correlation improves by 2\% and the mean correlation by 3.5\%.
For the contact PPG sensors on the forehead/ finger, the performance decreases again.
The median correlation decrease by 3.5\%/ 0\% and the mean correlation decreases by 7\%/ 5\%.

\subsection{Inferring EDA from Participants Without Significant EDA Responses}

In \autoref{tab:RGB_overall_results_bad}, we report the correlations between the inferred EDA signals using the total blood volume change, the blood pulsation amplitude, the instantaneous pulse rate, and the ground truth EDA signals of the four participants 01, 02, 12, and 14 that did not have a significant change in their ground truth EDA signals.
We can see that the obtained correlations drop considerably compared to the correlations obtained from the 12 participants with significant changes in the ground truth EDA signal.
Using the total blood volume change method, we obtain the highest median correlation from the videos of the participants' maxillary region ($\rho = 0.30$).
The median and mean correlations for all other regions except the palm and lips are below 0.20.
The results are also considerably lower using the contact PPG sensors, with a median correlation of $\rho = 0.18$ using the PPG sensor on the forehead and a median correlation of $\rho = 0.04$ using the PPG sensor on the finger.
For the blood pulsation amplitude the median correlations are similar to the total blood volume change with a maximum median correlation on the forehead ($\rho = 0.30$).
Using the instantaneous pulse rate, the correlations between the obtained signals and the ground truth EDA signals are close to zero for all regions and sensors.

\begin{table}[H]
    \centering
    \begin{tabular}{llcccccc}
    \toprule
         \textbf{Sensor} & \textbf{Region} & \multicolumn{2}{c}{\textbf{Blood vol. change}} & \multicolumn{2}{c}{\textbf{Blood puls. ampl.}} & \multicolumn{2}{c}{\textbf{Inst. pulse rate}} \\
         & & Median & Mean & Median & Mean & Median & Mean\\
         \midrule 
         \multirow{9}{*}{Camera} & Maxillary & \textbf{0.30} & \textbf{0.25} & 0.22 & 0.25 & -0.10 & -0.14\\
         & Palm & 0.26 & \textbf{0.25} & 0.14 & 0.06 & 0.01 & -0.07\\
         & Lips & 0.23 & 0.22 & 0.20 & 0.17 & -0.05 & -0.05\\
         & Whole face & 0.18 & 0.10 & 0.17 & 0.12 & 0.05 & \textbf{0.03} \\
         & Periorbital & 0.17 & 0.10 & 0.12 & 0.11 & -0.03 & -0.05\\
         & Nose & 0.14 & 0.14 & 0.17 & 0.12 & -0.05 & -0.11\\
         & Right cheek & 0.12 & 0.17 & 0.28 & 0.24 & -0.05 & -0.11\\
         & Left cheek & 0.11 & 0.14 & 0.22 & 0.21 & \textbf{0.06} & -0.06\\
         & Forehead & 0.0 & 0.02 & \textbf{0.30} & \textbf{0.33} & 0.04 & 0.0\\
         &&&&&&& \\
         \multirow{2}{*}{PPG sensor} & Forehead & \textbf{0.18} & \textbf{0.21} & \textbf{0.20} & \textbf{0.15} & -0.01 & -0.01\\
         & Finger & 0.04 & 0.12 & 0.02 & 0.1 & \textbf{0.04} & \textbf{0.04}\\
         \bottomrule
    \end{tabular}
    \caption{\textbf{Spearman correlations between the ground truth EDA signals and the predicted signals for the participants without significant EDA changes.}
    This shows the median and mean Spearman correlations between the contact EDA and the calculated signals from all regions and sensors using the total blood volume change, the mean blood pulsation amplitude, and the instantaneous pulse rate method.
    The values are obtained using only the four participants without significant changes in their ground truth EDA.
    The results are ordered in descending order for the median correlations obtained using the total blood volume changes.}
    \label{tab:RGB_overall_results_bad}
\end{table}

\subsection{Predicting Sympathetic Arousal Levels}
\label{sec:results_sympathetic_arousal_levels}
In \autoref{tab:arousal_levels}, we report the correlation values using Kendall's $\tau$ for predicting the different sympathetic arousal levels for the two best performing orofacial regions found in \hyperref[sec:region_interest]{Section 3.2.2.} and the two contact PPG sensors.
The correlation values across all regions and sensors are very similar with correlations between 0.73 and 0.83 (all correlations $\textit{p} < 0.01$).
The highest correlation is obtained from the videos of the forehead using the mean blood pulsation amplitude method with a median correlation of $\rho = 0.83$ (all correlations $\textit{p} < 0.01$).

\begin{table}[H]
    \centering
    \begin{tabular}{llcccccc}
    \toprule
         \textbf{Sensor} & \textbf{Region} & \multicolumn{2}{c}{\textbf{Blood vol. change}} & \multicolumn{2}{c}{\textbf{Blood puls. ampl.}} & \multicolumn{2}{c}{\textbf{Inst. pulse rate}}\\
         & & Median & Mean & Median & Mean & Median & Mean\\
         \midrule 
         \multirow{2}{*}{Camera} & Forehead & \textbf{0.80} & \textbf{0.78} & \textbf{0.83} & \textbf{0.81} & \textbf{0.78} & \textbf{0.75}\\
         & Palm & 0.80 & 0.77 & 0.70 & 0.68 & 0.78 & 0.73\\
         \multirow{2}{*}{PPG sensor} & Forehead & \textbf{0.81} & \textbf{0.80} & \textbf{0.82} & \textbf{0.77} & \textbf{0.79} & 0.74\\
         & Finger & \textbf{0.81} & 0.77& 0.80 & \textbf{0.77} & \textbf{0.79} & \textbf{0.75} \\
         \bottomrule
    \end{tabular}
    \caption{\textbf{Median and mean weighted Kendall's $\tau$ correlation between the ground truth EDA signals and the predicted signals.}
    This table shows the median and mean weighted Kendall's $\tau$ correlation across all participants between the contact EDA and the calculated signals for the forehead, palm, and finger regions using the total blood volume change, the mean blood pulsation amplitude, and the instantaneous pulse rate method.}
    \label{tab:arousal_levels}
\end{table}

\subsection{Motion Analysis}
\label{sec:results_motion_analysis}
In \autoref{tab:RGB_results_optical_flow}, we show the median and mean correlation values between the dense optical flow and the predicted signals from all three methods for the videos of the forehead and the palm, which showed the best performance in \hyperref[sec:region_interest]{Section 3.2.2.} among all orofacial regions.
As the optical flow can only be calculated for videos, no correlation values are reported for the contact PPG sensors.
All calculated correlation values are close to zero with the highest median correlation value being $\rho = 0.18$ from the video of the forehead using the mean blood pulsation amplitude.
The lowest correlation value is $\rho = -0.1$ from the video of the palm using the mean blood pulsation amplitude and the instantaneous pulse rate.

\begin{table}[H]
    \centering
    \begin{tabular}{llcccccc}
    \toprule
         \textbf{Sensor} & \textbf{Region} & \multicolumn{2}{c}{\textbf{Blood vol. change}} & \multicolumn{2}{c}{\textbf{Blood puls. ampl.}} & \multicolumn{2}{c}{\textbf{Inst. pulse rate}}\\
         & & Median & Mean & Median & Mean & Median & Mean\\
         \midrule 
         \multirow{2}{*}{Camera} & Forehead & \textbf{0.13} & \textbf{0.21} & \textbf{0.18} & -0.03 & \textbf{0.04} & \textbf{0.04} \\
         & Palm & 0.07 & -0.02 & 0.12 & \textbf{0.11} & -0.1 & -0.1 \\
         \bottomrule
    \end{tabular}
    \caption{\textbf{Median and mean Spearman correlations between the predicted signals and the dense optical flow.}
    This table shows the median and mean Spearman correlations between the dense optical flow and the calculated signals for all orofacial regions using the total blood volume change, the mean blood pulsation amplitude, and the instantaneous pulse rate method.}
    \label{tab:RGB_results_optical_flow}
\end{table}

\section{Discussion}
\subsection{How Well Can a Person's Sympathetic Arousal be Measured Using a Camera or a PPG Sensor?}
\label{sec:howwellmeasure}

We have demonstrated with our study that under ideal conditions, there is a correlation between a person's EDA and the change in intensity of the pixel values of a video of a person's face or hand. 
We have also verified that similar information can be derived from a contact PPG sensor on the forehead or finger of a person.
As such, the optically measured signal gives us an estimation of a person's sympathetic arousal that explains a majority of the variance in EDA.

In our quantitative analysis, we have analyzed all three of our methods (the total blood volume change, the blood pulsation amplitude, and the instantaneous pulse rate) and all four input signals (video of the face, video of the palm, contact PPG sensor on the finger, contact PPG sensor on the forehead).
Using the total blood volume change method, we achieve the highest median correlation across all participants from the video of the forehead ($\rho = 0.63$) and the video of the palm ($\rho = 0.57$) with a maximum individual correlation of 0.82 from the video of the forehead and a maximum individual correlation of 0.72 from the palm.
With the contact PPG sensors, we have obtained very similar median correlations from the recordings on the forehead ($\rho = 0.60$) and the recordings on the finger ($\rho = 0.59$) with a maximum individual correlation of 0.83 from the finger and 0.81 from the forehead.
Using the mean blood pulsation amplitude, which previous work shows correlates with blood perfusion~\cite{kamshilin2022bloodperfusion}, we achieve similar results than with the total blood volume change.
We achieve a maximum median correlation also on the forehead ($\rho = 0.58$) with a maximum individual correlation of 0.89 from a video of the forehead.
This indicates to us that we actually measure blood flow changes with our camera measurements as we discuss in the subsection below.
Using our third method, the instantaneous pulse rate, we obtained lower median correlations than using the total blood volume change or the mean blood pulsation amplitude.
From the video of the nose, we obtained the highest median correlation ($\rho = 0.49$), followed by the contact PPG sensor on the finger ($\rho = 0.47$), and the videos of the left ($\rho = 0.47$) and right cheek ($\rho = 0.46$).
Ultimately, it is the goal of our approaches to be able to predict a person's sympathetic arousal level to e.g., be able to detect when a person is in a state of high sympathetic arousal.
Therefore, we have also quantitatively analyzed how well our approaches can differentiate between low, medium, and high states of sympathetic arousal.
We have shown that all three of our approaches can differentiate between low, medium, and high arousal states using only the video of a person's face with high correlation values.
We obtain a median correlation of 0.83 using the mean blood pulsation amplitude, 0.8 using the total blood volume change, and 0.78 using the instantaneous pulse rate.

To visually evaluate the results, we show in \autoref{fig:example_plots_good} the obtained signals from all three methods compared to the ground truth EDA signals for four different participants and the four different regions which we found in our quantitative analysis are best to infer a person's sympathetic arousal.
We can see that the inferred signals from all three of our methods closely follow the overall tonic EDA signal.
Only small phasic changes are not captured.
However, we can also see that, while all three of our methods capture the overall trend of the ground truth EDA signal, our calculated signals and the ground truth EDA signals are not always perfectly synchronized.
For example, at the three minute mark of participant 07 in \autoref{fig:example_plots_good} the ground truth EDA signal is still decreasing while the predicted signals from the total blood volume change and the instantaneous pulse rate are already increasing.
This behavior is explained by two factors.
First, all three of our methods are non-causal filters as they use a sliding window approach (window size of 60 seconds for the total blood volume change and the mean blood pulsation amplitude, and a window size of between 24 and 84 seconds depending on the participant's pulse rate for the instantaneous pulse rate).
Our calculated signals, therefore, depend on future values from the input signals which can cause them to behave asynchronously to the ground truth EDA signals and increase before the ground truth EDA signals increase.
Second, we do not measure the ground truth EDA signals itself but physiological signals (blood volume change, blood pulsation amplitude, and instantaneous pulse rate) that correlate with the ground truth EDA signals.
As different physiological signals have different response times, our predicted signals and the ground truth EDA signals will not be perfectly synchronized.
Previous work has shown that the blood flow to the forehead is increased for a longer time period than the EDA on the forehead as a response to a painful gustatory stimulation~\cite{drummond1995mechanisms}.
Therefore, it is possible that our predicted signals decrease after the ground truth EDA signals decrease as we can see it, for example, for participant 07 in \autoref{fig:example_plots_good} for the mean blood pulsation amplitude after the second peak.

While a maximum median correlation of 0.63 from the videos of the foreheads is only a moderate correlation, related experiments have often reported correlations between 0.6–0.7 since physiological signals considerably vary across participants.
For example, Pai et al.~\cite{pai21hrvcam} report a correlation of 0.69 in two scenarios when predicting heart rate variability (HRV) from a video of the face.
While these two scenarios were under motion compared to our static dataset, Pai et al. estimated the same signal modality (HRV from camera vs. HRV from PPG contact sensor). 
In our work, we correlate two different signal modalities and aim to infer the sympathetic arousal (EDA from contact electrode) from peripheral hemodynamics (via camera).
All three of our approaches are able to to capture the overall tonic EDA signal, which is regarded a suitable measure of the activity of the sympathetic branch in stress states~\cite{boucsein2012eda, poh2010toniceda}, and can differentiate with high correlations between different sympathetic arousal levels.
Therefore, we believe that our introduced approaches offer a novel, unobtrusive way of measuring a person's sympathetic arousal using only an optical sensor such as a regular webcam.

\subsection{What Physiological Changes are We Measuring?}
\label{sec:whatwemeasure}

We can show, as described above, that we obtain a signal that strongly correlates with a person's EDA from the intensity pixel values of a video or the signal from a contact PPG sensor.
Although we cannot determine with certainty what underlying physiological mechanism is being captured, our results strongly suggest that we are measuring changes in blood flow, as we explain below.

Prior work has shown that through electrical stimulation, not only sweat gland activity but also blood flow to the forehead and the left cheek increases, and the blood flow to the finger decreases during arousing experiences~\cite{nordin1990forehead, vassend2005electropainblood}.
Measurement in the far infrared spectrum have revealed transient responses correlating with EDA that increase during stress in the periorbital, forehead, and maxillary regions, with the strongest response observed in the maxillary region~\cite{shastri2009edaface, shastri2012perinalsastress}. 
These measurements of temperature via a thermal camera captured heat dissipating as a result of sweat gland activity.
However, we obtain the highest correlations using the video of the forehead (see  \autoref{tab:RGB_overall_results}).
If we would be measuring sweat and not blood flow, the correlations from the maxillary region should be higher than the correlations from the forehead.


Furthermore, we show in \autoref{tab:RGB_overall_results} that the correlations using the contact PPG sensors are on par with the correlations using the videos. 
As variations in blood flow volume influence the signal from a contact PPG sensor~\cite{huiwen2022bloodflow}, this indicates to us that we actually measure blood flow.
Askarian et al.~\cite{askarian2019ppgwater} have shown that underwater, the signal amplitude of a gold-standard contact PPG sensor decreases by 22\,\%.
It could, therefore, be possible that the measurements from the contact PPG sensors are influenced by sweat.
But, as it would require considerable amounts of sweat to produce similar conditions as a contact PPG sensor held underwater, we believe that blood flow changes have a significantly bigger influence on the measured PPG signals than sweat.

In addition, previous work used the mean blood pulsation amplitude to calculate tissue blood perfusion in organs~\cite{kamshilin2022bloodperfusion}.
As we obtain very similar correlations with the total blood volume change method than with the mean blood pulsation amplitude, this indicates to us that we actually measure a change in blood flow that correlates with the ground truth EDA signal.

Finally, it is valid concern that our method does not capture a change in blood flow or sweat but other behaviors such as micro-expressions on the face or a change of the respiratory rate.
To address this issue, we filter out as many other influences to the input signals as possible by applying a 2\textsuperscript{nd} order Butterworth low pass filter with a cutoff frequency of 0.05\,Hz before we calculate the total blood volume change.
By allowing only frequencies below 0.05\,Hz, we can minimize the influence on our calculated signal from motion such as macro-expressions (duration between 0.5 and 4.0\,seconds~\cite{ekman2004emotions, matsumoto2011evidence}) or micro-expressions (duration between 0.067 and 0.5\,seconds~\cite{yan2013microexpressions, matsumoto2011evidence}), a person's heart rate (frequencies higher than 0.7\,Hz corresponding to 42\,bpm~\cite{spodick1992operational, heartrateaha}), a person's normal respiratory rate (0.2 to 0.33\,Hz~\cite{flenady2017respiratoryrate}), and non-linear dynamics of human respiration or Mayer waves (frequencies of 0.1\,Hz or higher~\cite{wysocki2006nonlinearbreathing, claude2006mayerwaves}).
Furthermore, while the videos of the participants' faces might be susceptible to, e.g. micro-expressions, the recorded PPG signals from the contact PPG sensors on the finger and forehead are considerably less susceptible to such movements.
Especially the contact PPG sensor on the finger is tightly mounted on the finger and would be less influenced by movements of the finger.
If our methods would only capture changes due to, e.g., micro-expression, we should see a difference in the correlations between the different input signals.
We do, however, obtain very similar correlation values and signals for all four modalities (video of the forehead, video of the palm, contact PPG sensor on the finger, contact PPG sensor on the forehead).
To also quantitatively assess that the motion in the face, e.g., due to facial expressions, does not influence our predicted signals, we calculated the dense optical flow for the videos of the forehead and the palm and compared it with our predicted signals.
The median correlations for both regions and all three proposed methods are between -0.1 and 0.21, which shows that our predicted signals do not correlate with the motion on the face.

Based on these results, we come to the conclusion that we predominantly measure a change in blood flow by calculating the absolute change and the mean blood pulsation amplitude, but we recognize that there may also be a minor contribution from sweat or motion.

\subsection{Limitations of Our Study and Methods}
\label{sec:limitations}

We see four limitations of our study and methods to measure a person's sympathetic arousal from a video or a PPG sensor.

First, we found that we could only infer tonic changes in the EDA signal. 
Neither of our three methods captured smaller, phasic changes reliably in the ground truth EDA signal (see \autoref{fig:example_plots_good}).
We hope that our work spurs future research into different methods that can also measure more subtle sympathetic responses.

Second, in order to obtain precise measurements from all sensors our study was highly controlled. 
Our strict protocol only allowed for evaluation of one stressor (pinching on the back of the arm or leg) that involved minimal body movements and did not trigger thermoregulation processes in the body.
As previous research suggests that different types of stressors trigger distinct cardiovascular, behavioral, and antinociceptive responses~\cite{janig1995pains}, further studies are needed to evaluate how well our methods work with different stressors.
We placed the participants' heads on a chin rest, their fingers under a belt, and kept the lighting and temperature in the room constant throughout the entire study.
In this way, we reduced all motions and external influences to a minimum to obtain input and ground truth signals that are as clean as possible.
Different methods might be needed for less constrained circumstances or real-world environments.
In addition, we did not measure the ground truth blood flow to the participants' faces and palms during the study.
We can, therefore, as discussed above, not determine with certainty that we actually measure a change in blood with our proposed method.

Furthermore, our dataset comprises data from participants from whom, according to the Fitzpatrick scale~\cite{fitzpatrick1988validity}, 5 individuals have skin type \RNum{2}, 11 skin type \RNum{3}, 3 skin type \RNum{5}, and 2 skin type \RNum{6}.
We acknowledge that a more balanced ratio of skin tones in our study would allow a deeper investigation about the performance of skin tone on the performance of our approaches as individuals with type \RNum{5} and \RNum{6} are under represented in our study.
We believe that this is an important topic to investigate as previous work has shown that camera-based rPPG prediction is more noise-prone for individuals with a larger melanin concentration~\cite{fallow2013influence, addison2018video, nowara2020meta}.
Further research is crucial, such as the recent work from Pai et al.~\cite{pai21hrvcam}, that improves the signal-to-noise ratio for motion scenarios and for individuals with larger melanin concentration.
In future work, we plan to conduct a study with a more balanced representation of different skin tones to evaluate how well our approaches work for different skin types.

Finally, we found that four out of 21 participants in our study did not have significant changes in their ground truth EDA as a response to pinching themselves.
Three of these four participants were also the participants with the lowest absolute EDA values (see \autoref{fig:preprocessing_eda}).
For these participants, all three of our proposed methods do not work with a maximum median correlation of 0.30 using the videos of the participants' maxillary areas and a median correlation of 0.20/ 0.04 using the contact PPG sensors on the participants' foreheads/ fingers (see \autoref{tab:RGB_overall_results_bad}).
Previous work observed similar results that for three out of 14 participants, there was no change in EDA as a response to electrical stimulation of the median nerve at the wrist~\cite{nordin1990forehead}.
The reason for this could either be the high inter-subject variability in EDA~\cite{posadoquintero2016edafrequency} or that some participants did not experience enough stress by pinching themselves to cause a sympathetic response.
It is, therefore, crucial to find a way in future work how we could determine with a camera or a PPG sensor if a person does not have significant changes in their EDA or sympathetic responses.

\section{Conclusion}
In conclusion, this paper demonstrates for the first time that it is possible to infer sympathetic arousal from a video of a person's face or hand using a regular RGB camera under normal lighting conditions or from a contact PPG sensor on a person's finger or forehead. 
By calculating the absolute change of the pixel intensity values or the mean blood pulsation amplitude of the rPPG/ contact PPG signals, we obtain median correlations of 0.58 to 0.63 between our inferred signals and the ground truth EDA signals with individual maximum correlations of up to 0.89.
We also show that all of our inferred signals closely follow the overall ground truth tonic EDA signals and achieve a median correlation of 0.83 for predicting if a person has a low, medium, or high sympathetic arousal level.
However, we also recognize that this approach so far only works in constrained settings where the movements of the participants are minimized and that our approach does not capture the phasic components of the ground truth EDA signals. 
By analyzing the different orofacial regions, we demonstrate the sympathetic arousal is best inferred from the forehead, finger, or palm as compared to the cheeks, nose or lips.
Furthermore, we found that four of the 21 participants in our study did not have significant changes in their EDA signal as a response to pinching themselves. 
For these participants, we were not able to infer their sympathetic arousal signal.

With this work, we have set an initial baseline for predicting sympathetic arousal from RGB videos or PPG sensors. 
This is a highly promising problem, as optical sensors such as regular RGB cameras or photodiodes are widely available on consumer electronics devices such as smartphones, smartwatches, and laptop computers. 
We believe that our work represents a significant contribution to the field of digital health by providing a non-invasive method for individuals to monitor their sympathetic arousal and stress responses. 
With the growing field of camera physiological measurement, we hope that knowledge from this domain could be utilized in the future to rapidly improve the measurement of sympathetic arousal from RGB videos or PPG sensors.

\section*{Funding}
This work was partially funded by the Swiss Joint Research Center from Microsoft.

\section*{Disclosures}
The authors declare no conflicts of interest.

\section*{Data availability} 
The data of the recorded dataset cannot be shared as the study participants did not agree to share their data, and the videos of the participants' faces cannot be de-identified.

\bibliography{sample}




\end{document}